%% file: main.tex
\documentclass[letterpaper,10pt,oneside]{article}

\usepackage{times}
\usepackage[margin=1.25in]{geometry}

\include{macros}

\include{notation}

\usepackage{algorithm}
\usepackage{algorithmic}
\graphicspath{{./fig/},{./fig_p3o/}}

\def \alg {DDPG++\xspace}

\title{
\alg: Striving for Simplicity in Continuous-control\\ Off-Policy Reinforcement Learning
}

\author{Rasool Fakoor$^1$, Pratik Chaudhari$^2$, Alexander J. Smola$^1$
\thanks{$^{1}$Amazon Web Services, $^{2}$University of Pennsylvania.
Correspondence to: fakoor@amazon.com
}}
\date{}

\begin{document}

\maketitle

\begin{abstract}
\noindent This paper prescribes a suite of techniques for off-policy Reinforcement Learning (RL) that simplify the training process and reduce the sample complexity. First, we show that simple Deterministic Policy Gradient works remarkably well as long as the overestimation bias is controlled. This is contrast to existing literature which creates sophisticated off-policy techniques. Second, we pinpoint training instabilities, typical of off-policy algorithms, to the greedy policy update step; existing solutions such as delayed policy updates do not mitigate this issue. Third, we show that ideas in the propensity estimation literature can be used to importance-sample transitions from the replay buffer and selectively update the policy to prevent deterioration of performance.
We make these claims using extensive experimentation on a set of challenging MuJoCo tasks. A short video of our results can be seen at \href{https://www.dropbox.com/sh/rpjua4wv6kzq7io/AAC5jnxbwc31840Y4jiXY9gGa?dl=0}{https://tinyurl.com/scs6p5m}.
\end{abstract}

\section{Introduction}
\label{s:intro}

Reinforcement Learning (RL) algorithms have demonstrated good performance on large-scale simulated data. It has proven challenging to translate this progress to real-world robotics problems for two reasons. First, the complexity and fragility of robots precludes extensive data collection. Second, a robot may face an environment during operation that is different than the simulated environment it was trained in; the myriad of intricate design and hyper-parameter choices made by current RL algorithms, especially off-policy methods, may not remain appropriate when the environment changes. We therefore lay out the following desiderata to make off-policy reinforcement learning-based methods viable for real-world robotics: (i) reduce the number of data required for learning, and (ii) simplify complex state-of-the-art off-policy RL algorithms.

\subsection{Problem Setup}

Consider a discrete-time dynamical system given by
\beq{
    \xkp = f(\xk, \uk, \xik)\ \textrm{given}\ x_0 \sim p_0
    \label{eq:f}
}
where $\xk, \xkp \in X \subset \reals^{n}$ are states at times $k$, $k+1$ respectively, $\uk \in U \subset \reals^p$ is the control input (also called action) applied at time $k$ and $\xik \in \reals^n$ is noise that denotes unmodeled parts of the dynamics. The initial state $x_0$ is drawn from some known probability distribution $p_0$.
We will work under the standard model-free formulation of RL wherein one assumes that the dynamics $f(\cdot, \cdot)$ is unknown to the learner. Consider the $\g$-discounted sum of rewards over an infinite time-horizon
\beq{
    v^{\uth}(x) =  \vth(x) = \E \SQBRAC{\sum_{k=0}^\infty \g^k r(\xk, \uk)\ |\ x_0 = x, \uth}.
    \label{eq:vth}
}
%u_k = \uth(x_k)\ \forall k
The left-hand side is known as the value function and the expectation is computed over trajectories of the dynamical system~\cref{eq:f}. Note that we always have $\xkp = f(\xk, \uth(\xk), \xik)$. The reward $r(\xk, \uk)$, denoted by $r_k$ in short, models
a user-chosen
incentive for taking the control input $u_k$ at state $x_k$. The goal in RL is to maximize the objective $J(\th) = \E_{x \sim p_0} \sqbrac{\vth(x)}$.

There are numerous approaches to solving the above problem. This paper focuses on off-policy learning  in which the algorithm reuses data from old polices to update the current policy~\cite{sutton2018reinforcement,bertsekas2019reinforcement}. The defining characteristic of these algorithms is that they use an experience replay buffer
\[
    D = \cbrac{(\xk,\uthp(\xk),r_k,\xkp}_{k=1,\ldots,N}
\]
collected by a policy $\uthp(x)$ to compute the value function $\vth(x)$ corresponding to the controller $\uth(x)$. In practice this replay buffer, $D$, consists of multiple trajectories
from different episodes. Off-policy techniques for continuous-state and control spaces regress a $\p$-parametrized action-value function $\qp(x,u)$ by minimizing the one-step squared-Bellman error (also called the temporal difference error)
\beq{
    \td^2(\p;\ \th) = \E_{(x,u,r,x') \in D} \sqbrac{\rbrac{\qp(x, u)- r - \g\ \qp(x', \uth(x'))}^2}.
    \label{eq:td}
}
If this objective is zero, the action-value function $\qp$ satisfies
\[
    \vth(x) = \qp(x, \uth(x)).
\]
which suggests that given $\qp$ one may find the best controller by maximizing
\beq{
    \ell(\th;\ \p) = \E_{(x,u,r,x') \in D} \SQBRAC{\qp(x, \uth(x))}.
    \label{eq:th}
}
The pair of equations~\cref{eq:td,eq:th} form a coupled pair of optimization problems with variables $(\th, \p)$ that can be solved by, say, taking gradient steps on each objective alternately while keeping one of the parameters fixed.

Although off-policy methods have shown promising performance in various tasks and are usually more sample efficient than on-policy methods~\cite{fakoor2019p3o, fujimoto2018addressing,haarnoja2018soft, Lillicrap2016ContinuousCW}, they are often very sensitive to hyper-parameters, exploration methods, among other things~\cite{HendersonRlMatter}.
This has led to a surge of interest in improving these methods.

\subsection{State of current algorithms}
\label{ss:sota}

The problems~\cref{eq:td,eq:th} form the basis for a popular off-policy method known as Deterministic Policy Gradient (DPG~\cite{silver2014deterministic}) or Deep-DPG~\cite{lillicrap2015continuous} which is its deep learning variant, as also many others such as Twin-Delayed DDPG (TD3~\cite{fujimoto2018addressing}) or Soft-Actor-Critic (SAC~\cite{haarnoja2018soft}). As written, this pair of optimization algorithms does not lead to good performance in practice. Current off-policy algorithms therefore introduce a number of modifications, some better motivated than others. These modifications have become \emph{de facto} in the current literature and we listen them below.
\begin{enumerate}[nosep]
    \item The $\td^2$ objective can be zero without $\qp(x,\uth(x))$ being a good approximation of the right hand side of~\cref{eq:vth}. Current algorithms use a ``target'' Q-function, e.g., they compute $\qp(x', \uth(x'))$ in~\cref{eq:td} using a time-lagged version of the parameters $\p$~\cite{mnih2013playing}. The controller $\uth$ in~\cref{eq:td} is also replaced by its time-lagged version
    \[
        \qp(x', \uth(x')) \la \qpt(x', \utht(x')).
    \]
    These target parameters $\p^t, \th^t$ are updated using exponential averages of $\p, \th$ respectively.

    \item The learnt estimate $\qp(x,u)$ typically over-estimates the right-hand side of~\cref{eq:vth}~\cite{Thrun1993}. TD3 and SAC therefore train two copies $\qp, \qpp$, and maintain different time-lagged targets $\qpot, \qptt$ for each to replace
    \[
        \qp(x', \uth(x')) \leftarrow \min \cbrac{\qpot(x', \uth(x')), \qptt(x', \uth(x'))}.
    \]
    This is called the ``double-Q'' trick~\cite{van2016deep}.

    \item Some algorithms like TD3 add ``target noise'' and use
    \[
        \qpot(x', \uth(x') + \textrm{noise}).
    \]
    while others such as SAC which train a stochastic controller $u' \sim \pith(\cdot\ |\ x)$ regularize with the entropy of the controller to get
    \[
        \qpot(x', u') - \a \log \pith(u'\ |\ x');
    \]
    here $\a > 0$ is a hyper-parameter.

    \item Further, SAC uses the minimum of two Q-functions $\qpone, \qptwo$ for the updating the controller in~\cref{eq:th} with the entropy term.

    \item The TD3 algorithm delays the updates to the controller, it performs two gradient-based updates of~\cref{eq:th} before updating the controller; this is called ``delaying policy updates''.
\end{enumerate}

\subsection{Contributions}
\label{ss:contributions}

Off-policy algorithms achieve good empirical performance on standard simulated benchmark problems using the above modifications. This performance comes at the cost of additional hyper-parameters and computational complexity for each of these modifications.

This paper presents a simplified off-policy algorithm named \alg that eliminates existing problematic modifications and introduces new ideas to make training more stable, while keeping the average returns unchanged. Our contributions are as follows:
\begin{enumerate}[nosep]
    \item We show that empirical performance is extremely sensitive to policy delay and there is no clear way to pick this hyper-parameter for all benchmark environments. We eliminate delayed updates in \alg.
    \item To make policy updates consistent with value function updates and avoid using over-estimated Q-values during policy updates, we propose to use minimum of two Q-functions. This part is the most critical step to make training stable.
    \item We observe that performance of the algorithm is highly dependent on the policy updates in~\cref{eq:th} because the estimate $\qp$ can be quite erroneous in spite of a small $\td^2$ in~\cref{eq:td}. We exploit this in the following way: observe that some tuples in the data $D$ depending upon the state $x$ may have controls that are similar to those of $\uth(x)$. These transitions are the most important to update the controller in~\cref{eq:th}, while the others in the dataset $D$ may lead to deterioration of the controller.
    We follow~\cite{fakoor2019meta, fakoor2019p3o} to estimate the propensity between the action distribution of the current policy and the action distribution of the past policies. This propensity is used to filter out transitions in the data that may lead to deterioration of the controller during the policy update.

    \item We show that adding noise while computing the target is not necessary for good empirical performance.
\end{enumerate}
Our paper presents a combination of a number of small, yet important, observations about the workings of current off-policy algorithms. The merit of these changes is that the overall performance on a large number of standard benchmark continuous-control problems is unchanged, both in terms of the rate of convergence and the average reward.

\section{\alg}
\label{s:approach}

We first discuss the concept of covariate shift from the machine learning and statistics literature in~\cref{ss:covariate_shift} and provide a simple method to compute it given samples from the dataset $D$ and the current policy being optimized $\uth(x)$ in~\cref{ss:logistic_regression}. We then present the algorithm in~\cref{ss:algorithm}.

\subsection{Covariate shift correction }
\label{ss:covariate_shift}

Consider the supervised learning problem where we observe independent and identically distributed data from a distribution $q(x)$, say the training dataset. We would however like to minimize the loss on data from another distribution $p(x)$, say the test data. This amounts to minimizing
\beq{
    \aed{
        \E_{x \sim p(x)}\ &\E_{y|x}\ \sqbrac{\ell(y, \hat{y}(x))}\\
        &= \E_{x\sim q(x)}\ \E_{y|x}\ \sqbrac{\b(x)\ \ell(y, \hat{y}(x))}.
    }
    \label{eq:is}
}
Here $y$ are the labels associated to draws $x \sim q(x)$ and $\ell(y, \hat{y}(x))$ is the loss of the predictor $\hat{y}(x)$.
The importance ratio is defined as
\beq{
    \b(x) := \f{\d p(x)}{\d q(x)}
    \label{eq:w}
}
which is the Radon-Nikodym derivative of the two densities~\cite{resnick2013probability} and it re-balances the data to put more weight on unlikely samples in $q(x)$ that are likely under the test data $p(x)$. If the two distributions are the same, the importance ratio is 1 and doing such correction is unnecessary. When the two distributions are not the same, we have an instance of covariate shift and need to use the trick in~\cref{eq:is}.

\subsection{Logistic regression for estimating the covariate shift}
\label{ss:logistic_regression}

When we do not know the densities $q(x)$ and $p(x)$ and we need to estimate $\b(x)$ using some finite data $X_q = \cbrac{x_1, \ldots, x_m}$ drawn from $q$ and $X_p = \cbrac{x_1', \ldots, x_m'}$ drawn from $p$. As~\cite{agarwal2011linear} show, this is easy to do using logistic regression. Set $z_k = 1$ to be the labels for the data in $X_q$ and $z_k = -1$ to be the labels of the data in $X_p$ for $k \leq m$ and fit a logistic classifier on the combined $2m$ samples by solving
\beq{
    w^* = \min_{w}\ \f{1}{2m} \sum_{(x,z)}\ \log \rbrac{1 + e^{-z w^\top x}} + c\ \norm{w}^2.
    \label{eq:logistic}
}
This gives
\beq{
    \b(x) = \f{\P(z = -1\ |\ x)}{\P(z = 1\ |\ x)} = e^{-{w^*}^\top x}.
    \label{eq:beta_exp_f}
}
This method of computing propensity score, or the importance ratio, is close to two-sample tests~\cite{fakoor2019p3o, agarwal2011linear, Reddi2015DoublyRC} in the statistics literature.

\begin{remark}[Replay buffer has diverse data]
The dataset in off-policy RL algorithms, also called the replay buffer, is created incrementally using a series of feedback controllers obtained during interaction with simulator or the environment. This is done by drawing data from the environment using the initialized controller $u_{\th_0}$; the controller is then updated by iterating upon~\cref{eq:td,eq:th}. More data is drawn from the new controller and added to the dataset $D$. The dataset for off-policy algorithms therefore consists of data from a large number of \emph{diverse} controllers/policies. It is critical to take this diversity into consideration when sampling from the replay buffer; propensity estimation allows doing so.
\end{remark}

\subsection{Algorithm}
\label{ss:algorithm}

This section elaborates upon the \alg algorithm along some implementation details that are important in practice.

\begin{figure}[!htpb]
\centering
\captionsetup[subfigure]{justification=centering}
\begin{subfigure}[t]{0.4\textwidth}
    \centering
    \includegraphics[width=\textwidth]{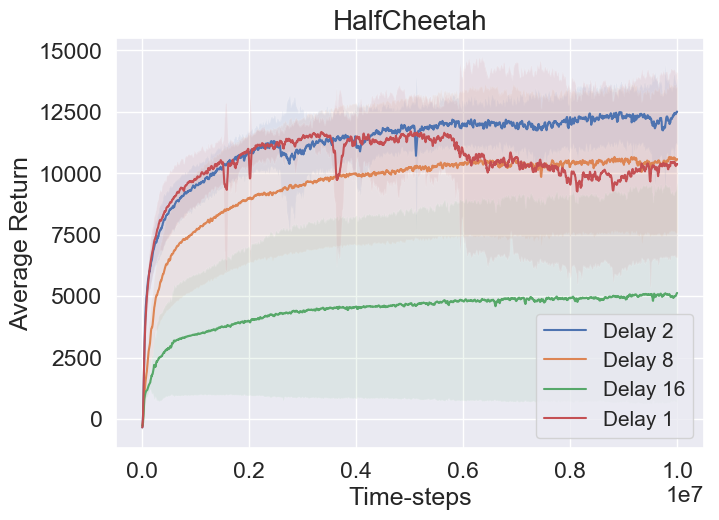}
    \caption{}
    \label{fig:HalfCheetah__sens}
\end{subfigure}
\begin{subfigure}[t]{0.4\textwidth}
    \centering
    \includegraphics[width=\textwidth]{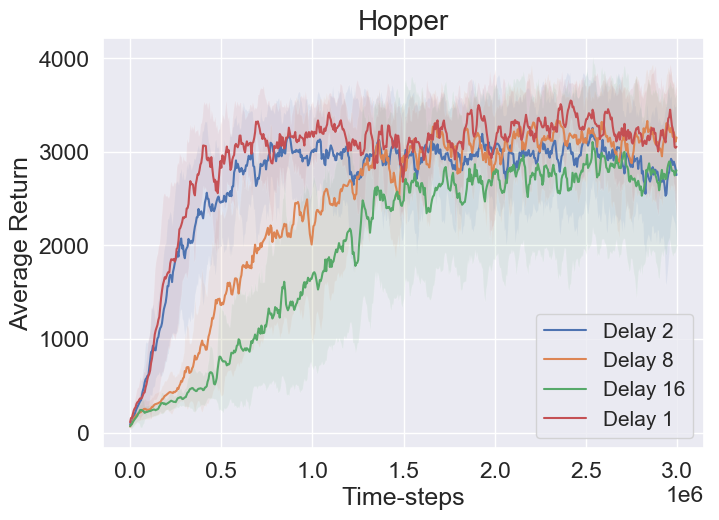}
    \caption{}
    \label{fig:Hopper__sens}
\end{subfigure}
\caption{\tbf{Performance is sensitive to policy delay hyper-parameters}. \cref{fig:HalfCheetah__sens} shows the average return over evaluation episodes on the HalfCheetah environment in MuJoCo for policy delay of 1 , 2, 8 and 16 steps. The performance with a delay of 2, where the gradient-update frequency of~\cref{eq:th} is half that of~\cref{eq:td} is better than the others. However, this trend is not consistent with that in the Hopper environment where all delays perform about the same with minor differences in how fast learning occurs.}
\label{fig:delay}
\end{figure}

\noindent \tbf{Policy delay causes instability.}
\cref{fig:HalfCheetah__sens} shows that different policy delays lead to large differences in the performance for the HalfCheetah environment of OpenAI's Gym~\cite{brockman2016openai} in the MuJoCo~\cite{todorov2012mujoco} simulator. The second figure, in~\cref{fig:Hopper__sens}, shows that different policy delays perform about the same for the Hopper environment; the performance for the other MuJoCo environments is the same as that of the Hopper. Policy delay was introduced by the authors in~\cite{fujimoto2018addressing} to stabilize the performance of off-policy algorithms but this experiment indicates that policy delay may not be the correct mechanism to do so.

\begin{algorithm}[!t]
\small
\renewcommand{\arraystretch}{0.7}
  \caption{\alg algorithm}
  \label{alg:ddpgpp}
  \begin{algorithmic}
    \STATE Initialize neural networks $\qpone, \qptwo$ for the value function, their targets $\qpot, \qptt$, the controller $\uth$ and its corresponding target $\utht$.
    \STATE Initialize the dataset $D = \emptyset$
    \FOR{$i$ = 1,\ldots,$\f{N}{T}$}
        \STATE Sample a trajectory of the system $\xkp = f(\xk, \uk)$ where $\uk = \uth(\xk) + \textrm{noise}$ for exploration. Record the reward $r(\xk, \uk)$ at each timestep and add the tuple $(\xk, \uk, r(\xk, \uk), \xkp)$ for $k = 1, \ldots, T$ to the dataset $D$.

    \FOR{$t = 1,\ldots, T$}

        \STATE 1. Sample a mini-batch of tuples $(x, u, r, x') \in \BB \subset D$.

        \STATE 2. Compute target
            \[
                y = r + \g \min\cbrac{\qpot(x', \utht(x')), \qptt(x', \utht(x')).}
            \]

        \STATE 3. Update parameters of the value networks $\qpone, \qptwo$ using the gradient of the form
            \[
                \f{1}{\abs{\BB}} \grad_\p \sum_{(x,u,r,x') \in \BB} \rbrac{y - \qp(x,u)}^2.
            \]

        \STATE 4. If using propensity-based controller updates, estimate the importance ratio $\b(x)$ (see~\cref{ss:algorithm}) by fitting a logistic classifier to discriminate between the sets of controls $\cbrac{u:\ (x,u,r,x') \in \BB}$ and $\cbrac{\uth(x):\ (x,u,r,x') \in \BB}$. Normalize
        \[
            \tilde{\b}(x) = \f{\b(x) - \min_x \b(x)}{\max_x \b(x) - \min_x \b(x)}
        \]
        where $\min, \max$ are computed over all states $x$ such that $(x,u,r,x') \in \BB$.
        Set $\tilde{\b}(x) = 1$ for all $x$ if propensity updates are not being used.

        \STATE 5. Update the controller $\uth$ using the gradient
            \[
                \f{1}{\abs{\BB}} \grad_\th \sum_{(x,u,r,x') \in \BB} \tilde{\b}(x)\ \min\rbrac{\qpone(x, \uth(x)), \qptwo(x, \uth(x))}.
            \]

        \STATE 6. Exponential averaging to update the target
            \[
                \aed{
                    \p_1^t &\la (1-\t)\ \p_1^t + \t \p_1\\
                    \p_2^t &\la (1-\t)\ \p_2^t + \t \p_2\\
                    \th^t &\la (1-\t)\ \th^t + \t \th.
                }
            \]
        \ENDFOR
    \ENDFOR
  \end{algorithmic}
\end{algorithm}

\noindent \tbf{Propensity estimation for controls.}
The controller $\uth$ is being optimized to be greedy with respect to the current estimate of the value function $\qp$. If the estimate of the value function is erroneous--- theoretically $\qp$ may have a large bias because its objective~\cref{eq:td} only uses the one-step Bellman error---updates to the controller will also incur a large error. One way to prevent such deterioration is to update the policy only on states where the control in the dataset $U = \cbrac{(u,1):\ (x,u,r,x') \in D}$ and the current controller's output $U' = \cbrac{(\uth(x),-1): (x,u,r,x') \in D}$ are similar for a given state $x$. The idea is that since the value function is fitted across multiple gradient updates of~\cref{eq:td} using the dataset, the estimate of $\qp$ should be consistent for these states and controls. The controller $\uth$ is therefore being evaluated and updated only at states where the $\qp$ is consistent.

It is easy to instantiate the above idea using propensity estimation in~\cref{ss:logistic_regression}. At each iteration before updating the policy using~\cref{eq:th} we fit a logistic classifier on the two datasets $U$ and $U'$ above to estimate the likelihood $\f{\P(z = -1\ | x)}{\P(z = 1\ | x)}$ which is the relative probability of a control $u$ coming from a dataset $U'$ versus $U$. The objective for the policy update in~\cref{eq:th} is thus modified to simply be
\beq{
    \E_{(x,u,r,x') \in D} \SQBRAC{\b(x)\ \qp(x, \uth(x))}.
    \label{eq:th_beta}
}
Note that higher the $\b$, closer the control in the dataset to the output of the current controller $\uth$. Also observe that the ideal accuracy of logistic classifier for propensity estimation is 0.5, i.e., the classifier should not be able to differentiate between controls in the dataset and those taken by the controller. As a result, $\b$ will have large constant values. The importance ratio $\b$ will be small if the current controller's output at the same state $x$ is very different; this modified objective discards such states while updating $\uth$.

\begin{figure}[!htpb]
\centering
\captionsetup[subfigure]{justification=centering}
\begin{subfigure}[t]{0.4\textwidth}
    \centering
    \includegraphics[width=\textwidth]{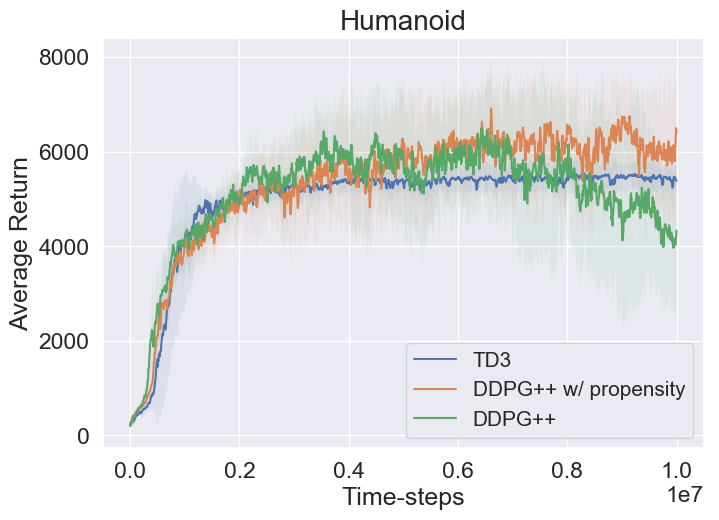}
    \caption{}
    \label{fig:Humanoid__prop}
\end{subfigure}
\begin{subfigure}[t]{0.4\textwidth}
    \centering
    \includegraphics[width=\textwidth]{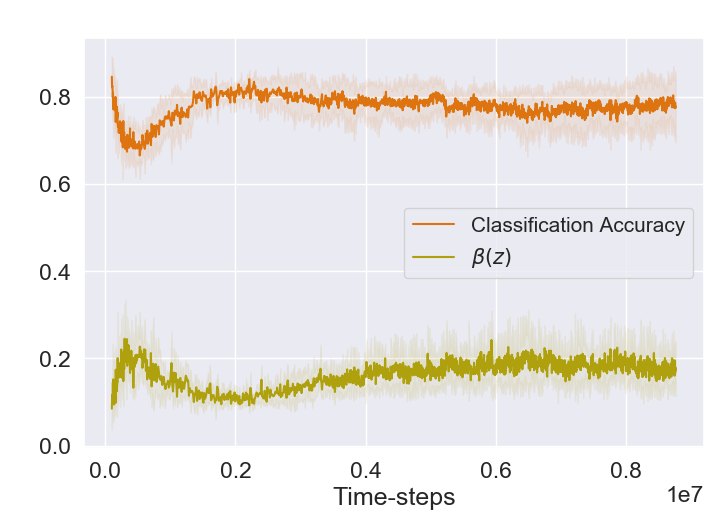}
    \caption{}
    \label{fig:Humanoid_beta_prox}
\end{subfigure}
\caption{\tbf{Effect of adding propensity to the policy updates}. ~\cref{fig:Humanoid__prop} shows the average training returns for TD3, \alg with propensity (orange) and \alg without propensity (green). Using \alg with propensity-based weighing leads to higher returns, on average, and does not suffer from the degradation seen in the green curve towards the end of training. \cref{fig:Humanoid_beta_prox} shows the classification accuracy of the logistic classifier (orange) and the average value of the propensity across mini-batches (yellow) during training. Humanoid is a high-dimensional control task and controls in the dataset are quite different from those the controller $\uth$ would take. The propensity $\b$ is \emph{constant} during training which suggests that even if the size of the dataset is growing, new controls added to the dataset are closer to $\uth$.}
\label{fig:propensity}
\end{figure}

\begin{remark}[Relation to Batch-Constrained-Q-learning]
\label{rem:relation_to_bcq}
Our modification to the objective in~\cref{eq:th_beta} is close to the idea of the BCQ algorithm in~\cite{fujimoto2018off}. This algorithm uses a generative model to learn the action distribution at new states in Q-learning to update $\qp$ in ~\cref{eq:td} selectively. The BCQ algorithm is a primarily imitation-based algorithm and is designed for the so-called offline reinforcement learning setting where the dataset $D$ is given and fixed, and the agent does not have the ability to collect new data from the environment. Our use of propensity for stabilizing the controller's updates in~\cref{eq:th_beta} is designed for the off-policy setting and is a simpler instantiation of the same idea that one must selectively perform the updates to $\qp$ and $\uth$ at specific states.
\end{remark}

We next demonstrate the performance of propensity-weighted controller updates in a challenging example in~\cref{fig:propensity}. \cref{fig:Humanoid__prop} compares the performance of TD3, \alg and \alg with propensity weighing. The average return of propensity weighing (orange) is higher than the others; the unweighted version (green) performs well but suffers from a sudden degradation after about 6M training samples. We can further understand this by observing~\cref{fig:Humanoid_beta_prox}: the propensity $\b(x)$ is always quite small for this environment which suggests that most controls in the dataset are very different from those the controller $\uth$ would take. Using propensity to weight the policy updates ignores such states and protects the controller from degradation seen in~\cref{fig:Humanoid__prop}.

\section{Experimental Validation}
\label{s:expt}

\noindent \tbf{Setup.} We use the MuJoCo~\cite{todorov2012mujoco} simulation environment and the OpenAI Gym~\cite{brockman2016openai} environments to demonstrate the performance of \alg. These are simulated robotic systems, e.g., the HalfCheetah environment in~\cref{fig:HalfCheetah__sens} has 17 states and 6 controls and is provided a reward at each time-step which encourages velocities of large magnitude and penalizes the magnitude of control. The most challenging environment in this suite is the Humanoid task which has 376-dimensional state-space and 17 control inputs; the Humanoid is initialized upright and is given a reward for moving away from its initial condition with quadratic penalty for control inputs and impact forces. All the hyper-parameters are given in the Appendix.

\noindent \tbf{Baseline algorithms.} We compare the performance of \alg against two algorithms. The first is TD3 of~\cite{fujimoto2018addressing} which is a recent off-policy algorithm that achieves good performance on these tasks. The second is DDPG~\cite{lillicrap2015continuous} which is, relatively speaking, a simple algorithm for off-policy learning. The \alg algorithm discussed in this paper can be thought of as a simplification of TD3 to bring it closer to DDPG. The SAC algorithm~\cite{haarnoja2018soft} has about the same performance as that of TD3; we therefore do not use it as a baseline.

\begin{figure*}
\centering
\captionsetup[subfigure]{justification=centering}
\begin{subfigure}[t]{0.32\textwidth}
\centering
\includegraphics[width=\textwidth]{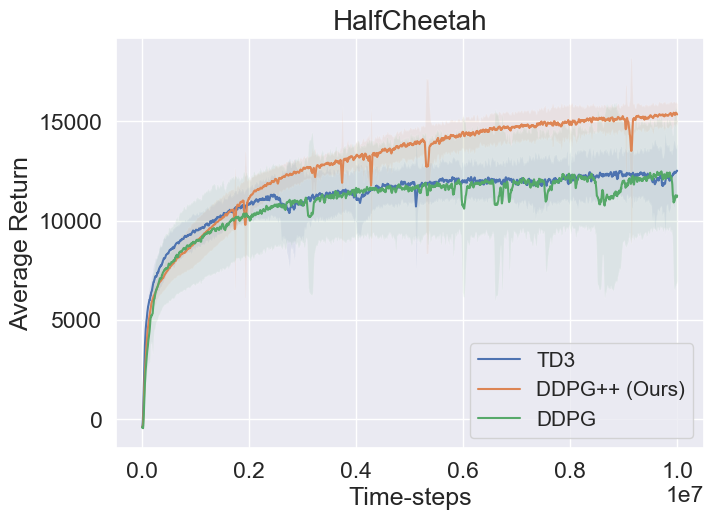}
\label{fig:HalfCheetah_mc_mj_}
\end{subfigure}%
~
\begin{subfigure}[t]{0.32\textwidth}
\centering
\includegraphics[width=\textwidth]{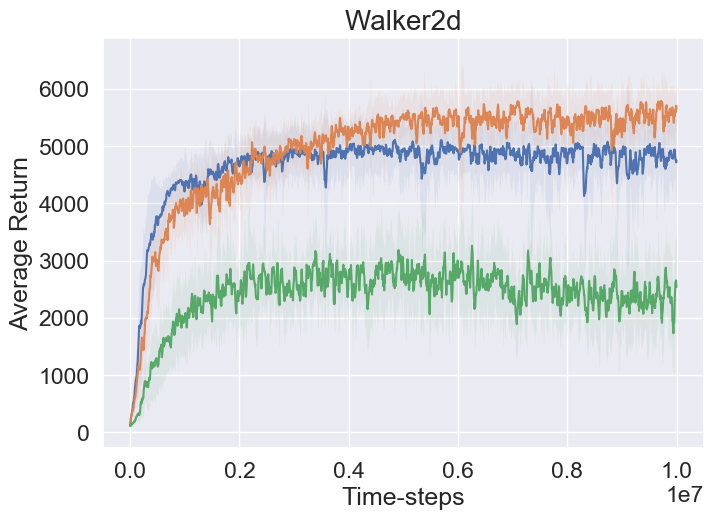}
\label{fig:Walker2d_mc_mj_}
\end{subfigure}%
~
\begin{subfigure}[t]{0.32\textwidth}
\centering
\includegraphics[width=\textwidth]{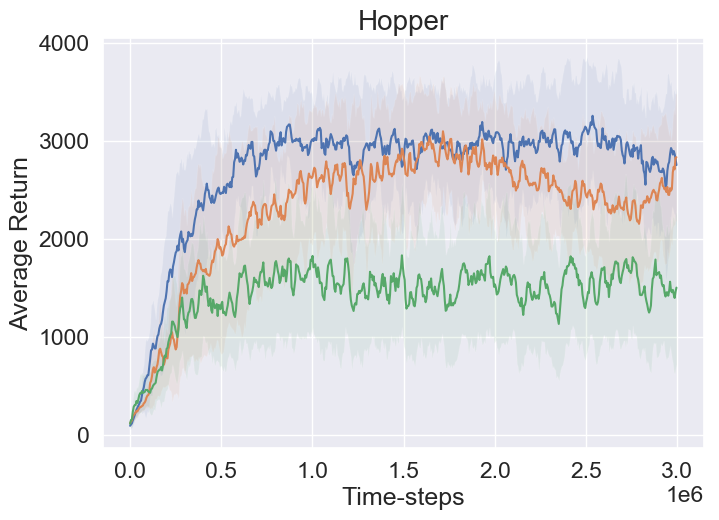}
\label{fig:Hopper_mc_mj_}
\end{subfigure}%

\begin{subfigure}[t]{0.32\textwidth}
\centering
\includegraphics[width=\textwidth]{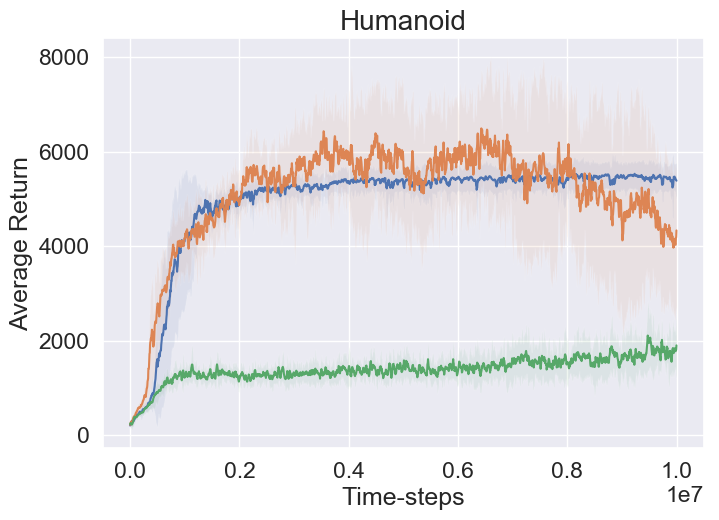}
\label{fig:Humanoid_mc_mj_}
\end{subfigure}%
~
\begin{subfigure}[t]{0.32\textwidth}
\centering
\includegraphics[width=\textwidth]{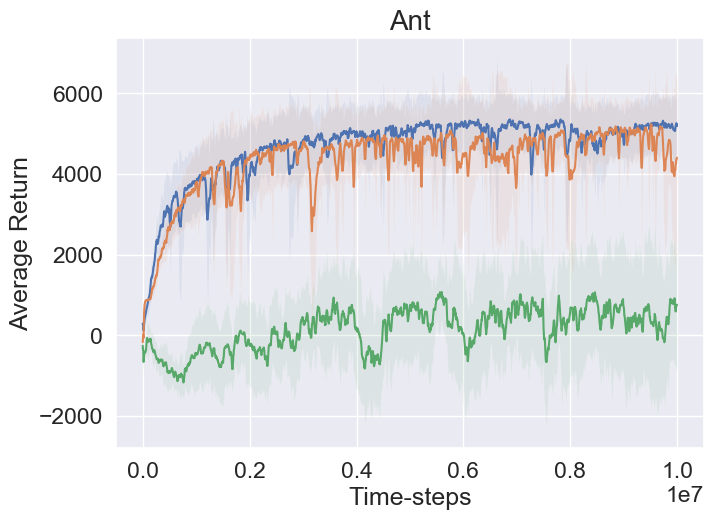}
\label{fig:Ant_mc_mj_}
\end{subfigure}%
~
\begin{subfigure}[t]{0.32\textwidth}
\centering
\includegraphics[width=\textwidth]{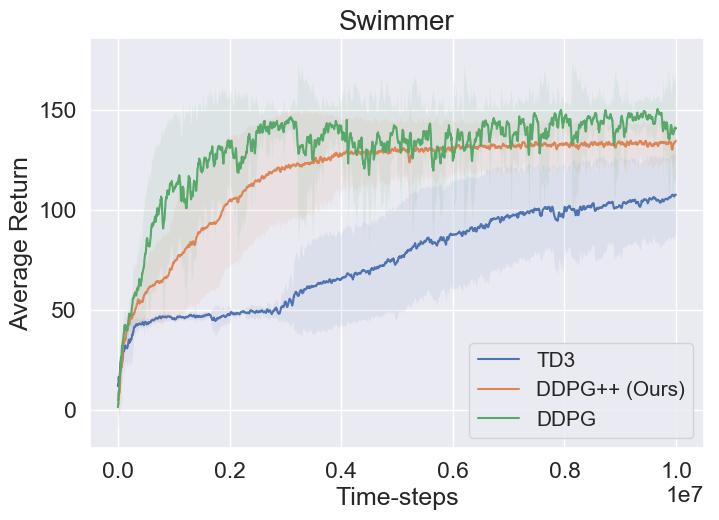}
\label{fig:Swimmer_mc_mj_}
\end{subfigure}%
~
\caption{\tbf{Comparison of DDPG (green), TD3 (blue) and \alg (orange) on standard continuous-control tasks in MuJoCo}. Across a suite of tasks, the simplified \alg algorithm performs comparably to TD3. This experiment shows that even if the DDPG algorithm is consistently worse that the off-policy algorithms that followed it, one can get good performance with a few simple additions discussed in this paper.}
\label{fig:mujoco}
\end{figure*}

\noindent \tbf{Performance on standard tasks.}
\cref{fig:mujoco} shows the average returns of these three algorithms on MuJoCo environments. Across the suite, the performance of \alg (without propensity correction) is stable and at least as good as TD3. Performance gains are the largest for HalfCheetah and Walker-2D. This study shows that one can obtain good performance on the MuJoCo suite with a few simple techniques while eliminating a lot of complexity of state-of-the-art algorithms. The average returns at the end of training are shown in~\cref{tab:mujoco}. The performance of \alg with propensity was about the same as that of \alg without it for the environments in~\cref{fig:mujoco}.

\cref{fig:propensity} shows that propensity-weighing is dramatically effective for the Humanoid environment. This is perhaps because of the significantly higher dimensionality of this environment as compared to others in the benchmark. Hyper-parameters tuned for other environments do not perform well for Humanoid, propensity-weighing the policy updates is much easier and more stable, in contrast.

\subsection{Dexterous hand manipulation}

\begin{figure}[!htpb]
\centering
\captionsetup[subfigure]{justification=centering}
\begin{subfigure}[t]{0.4\textwidth}
    \centering
    \includegraphics[width=\textwidth]{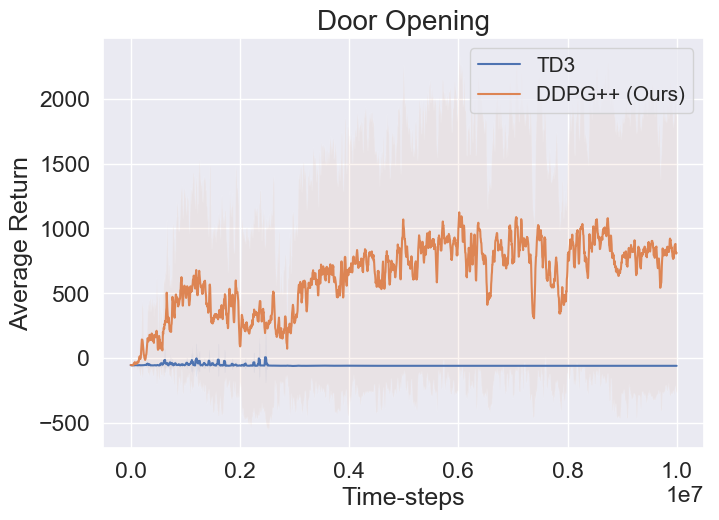}
    \caption{}
    \label{fig:door-v0_new_tasks}
\end{subfigure}
\hspace{0.2in}
\begin{subfigure}[t]{0.25\textwidth}
    \centering
    \includegraphics[width=\textwidth]{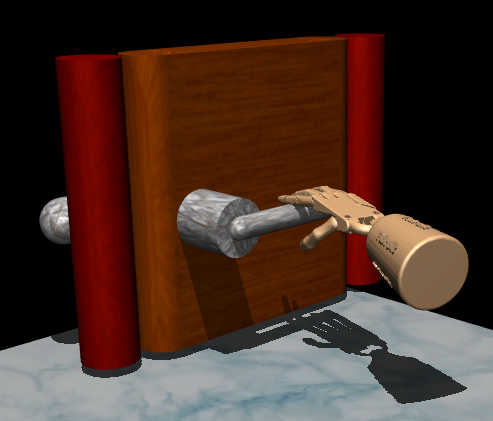}
    \caption{}
    \label{fig:door_opening_new_tasks}
\end{subfigure}
\caption{\tbf{Comparison of TD3 and \alg on the door opening task using a simulated ADROIT hand.}}
\label{fig:door}
\end{figure}

We next evaluate the \alg algorithm on a challenging robotic task where a simulated anthropomorphic ADROIT hand~\cite{kumar2013fast} with 24 degrees-of-freedom is used to open a door~\cref{fig:door_opening_new_tasks}. This task is difficult because of significant dry friction and a bias torque that forces the door to stay closed. This is a completely model-free task and we do not provide any information to the agent about the latch, the RL agent only gets a reward when the door touches the stopper at the other end. The authors in~\cite{rajeswaran2017learning} demonstrate that having access to expert trajectories is essential for performing this task. Reward shaping, where the agent's reward is artificially bolstered the closer to the goal as it gets, can be helpful when there is no expert data. The authors in ~\cite{rajeswaran2017learning} demonstrate that DDPG with observations (DDPGfD~\cite{vecerik2017leveraging}) performs poorly on this task.
\cref{fig:door-v0_new_tasks} shows the average evaluation returns of running \alg on this task without any expert data and~\cref{tab:scv_rate} shows the success rate as the algorithm trains. Hyper-parameters for this task are kept the same as those in continuous-control tasks in the MuJoCo suite.

\begin{table}[!htpb]
\renewcommand{\arraystretch}{1.1}
\begin{center}
\small
\caption{\tbf{Average success rate on ADROIT hand.}}
\label{tab:scv_rate}
{
\begin{tabular}{p{2cm} r r}
Task            & TD3       & \alg \\
\toprule
ADROIT hand  & 0 & \tbf{34}\\
%\bottomrule
\end{tabular}
}
\end{center}
\end{table}

\begin{table}[!htpb]
\renewcommand{\arraystretch}{1.3}
\begin{center}
\small
\caption{\tbf{Average return (10 different runs of the algorithm on different random seeds) on 1000 evaluation trajectories for some standard continuous-control tasks on MuJoCo}. 10M training samples were used for all environments except Hopper which used 3M.}
\vspace*{0.05in}
\label{tab:mujoco}
{
\begin{tabular}{p{2cm} r r r}
% \rowcolor{gray!15}
Task   & DDPG   & TD3       &  \alg  \\
\midrule
HalfCheetah  & 11262 & 12507 & \tbf{15342}\\
Walker2d  & 2538 & 4753 & \tbf{5701}\\
Hopper  & 1501 & 2752 & \tbf{2832}\\
Humanoid  & 1898 & \tbf{5384} & 4331\\
Ant  & 758 & \tbf{5185} & 4404\\
Swimmer  & \tbf{141} & 107 & 134\\
%\bottomrule
\end{tabular}
}
\end{center}
\end{table}

\begin{table}[!htpb]
\renewcommand{\arraystretch}{1.1}
\begin{center}
\small
\caption{\tbf{Average return on MuJoCo continuous-control tasks} on 10 seeds.}
\label{tab:mujoco}
{
\begin{tabular}{p{2cm} r r r}
% \rowcolor{gray!15}
Task            & TD3       & \alg      &  \alg\\
&&& (with propensity)\\
\toprule
Humanoid        & 5384      & 4331      & \tbf{6384}\\
%\bottomrule
\end{tabular}
}
\end{center}
\end{table}

\section{Discussion}
\label{s:discussion}

We discussed an algorithm named \alg which is a simplification of existing algorithms in off-policy reinforcement learning. This algorithm eliminates noise while computing the value function targets and delayed policy updates in standard off-policy learning algorithms; doing so eliminates hyper-parameters that are difficult to tune in practice. \alg uses propensity-based weights for the data in the mini-batch to update the policy only on states where the controls in the dataset are similar to those that the current policy would take. We showed that a vanilla off-policy method like DDPG works well as long as overestimation-bias in value estimation is reduced. We evaluated these ideas on a number of challenging, high-dimensional simulated control tasks.

Our paper is in the spirit of simplifying complex state-of-the-art RL algorithms and making them more amenable to experimentation and deployment in robotics. There are a number of recent papers with the same goal, e.g.,~\cite{fu2019diagnosing} is a thorough study of Q-learning algorithms,~\cite{fujimoto2018addressing} identifies the key concept of overestimation bias in Q-learning,~\cite{agarwal2019striving} shows that offline Q-learning, i.e., learning the optimal value function from a fixed dataset can be performed easily using existing algorithms if given access to a large amount of exploratory data.
While these are promising first results, there a number of steps that remain before reinforcement learning-based methods can be incorporated into standard robotics pipelines. The most critical issue that underlies all modern deep RL approaches is that the function class that is used to approximate the value function, namely the neural network, is poorly understood. Bellman updates may not be a contraction if the value function is approximated using a function-approximator and consequently the controls computed via the value function can be very sub-optimal. This is a fundamental open problem in reinforcement learning~\cite{antos2008fitted,antos2008learning,farahmand2010error,munos2005error} and solutions to it are necessary to enable deployment of deep RL techniques in robotics.

\bibliographystyle{unsrt}
\bibliography{dl,main,offline}

\appendix

\section{Appendix A: Hyper-parameters}
\label{s:hp}

We next list down all the hyper-parameters used for the experiments in~\cref{s:expt}.

\begin{table}[!htpb]
\renewcommand{\arraystretch}{1.2}
\small
\centering
\caption{\tbf{Hyper-parameters for DDPG, TD3, and DDPG++  for continuous-control benchmark tasks.} We use a network with two full-connected layers (256 hidden neurons each) for all environments. The abbreviations HC, AN, HM stand for Half-Cheetah, Ant and Humanoid.}
\label{tab:hp}
\begin{tabular}{p{3.1cm} r r r }
\toprule
Parameters & DDPG & TD3 & \alg (Ours)  \\
\midrule
Exploration noise       & 0.1 & 0.1 & 0.2 \\
Policy noise            & N/A & 0.2 & N/A \\
Policy update frequency & N/A & 2 & N/A\\
Adam learning rate     & 0.001 & 0.001 & 0.0003\\
Adam learning rate (HM)      & 1E-4 & 1E-4 & 1E-4\\

Hidden size            & 256 & 256 & 256\\
Burn-in          & 1000 & 1000 & 1000\\
Burn-in (HC \& AN)          & 10000 & 10000 & 10000\\
Batch size        & 100 & 100 & 100 \\
\bottomrule
\end{tabular}
\end{table}

\end{document}

%% file: macros.tex
\usepackage{amssymb, amsthm, bm, mathtools}
\usepackage[scr=boondoxo]{mathalfa}
\usepackage[usenames,dvipsnames,svgnames,table]{xcolor}
\usepackage[pdftex, xetex]{graphicx}
\usepackage[font={small}]{caption}
\usepackage{setspace}
\usepackage[shortlabels]{enumitem}
\usepackage{float, colortbl, tabularx, longtable, multirow, subcaption, environ, wrapfig, textcomp, booktabs}
\usepackage{pgf, tikz, framed}
\usepackage[normalem]{ulem}
\usetikzlibrary{arrows,positioning,automata,shadows,fit,shapes}
\usepackage[english]{babel}
\usepackage{blindtext}
\usepackage{hyperref,url,cite}

\hypersetup{
    colorlinks=true,
    citecolor=blue,
}

\usepackage{microtype}
\microtypecontext{spacing=nonfrench}

\makeatletter
\def\th@plain{%
  \thm@notefont{}% same as heading font
  \itshape % body font
}
\def\th@definition{%
  \thm@notefont{}% same as heading font
  \normalfont % body font
}
\makeatother

\usepackage[capitalize]{cleveref}
\crefformat{equation}{(#2#1#3)}
\crefmultiformat{equation}{(#2#1#3)}%
{ and~(#2#1#3)}{, (#2#1#3)}{ and~(#2#1#3)}
\crefformat{theorem}{Theorem~#2#1#3}
\crefformat{lemma}{Lemma~#2#1#3}
\crefformat{remark}{Remark~#2#1#3}
\crefformat{section}{Section~#2#1#3}
\crefformat{assumption}{Assumption~#2#1#3}
\crefformat{example}{Example~#2#1#3}
\crefrangeformat{section}{Sections~#3#1#4--#5#2#6}
\crefformat{assumption}{Assumption~#2#1#3}
\crefformat{example}{Example~#2#1#3}

\crefrangeformat{equation}{~(#3#1#4--#5#2#6)}
\crefrangeformat{figure}{Figures~#3#1#4--#5#2#6}
\crefrangeformat{assumption}{Assumptions~(#3#1#4--#5#2#6)}

\newcommand{\tbf}[1]{\textbf{#1}}
\DeclarePairedDelimiter{\abs}{\lvert}{\rvert}

\DeclarePairedDelimiter{\norm}{\lVert}{\rVert}

\newcommand{\aed}[1]{\begin{aligned} #1 \end{aligned}}
\newcommand{\beq}[1]{\begin{equation}#1\end{equation}}

\newcommand{\la}{\ \leftarrow\ }

\providecommand\f[2]{\ensuremath \frac{#1}{#2}}

\providecommand\f[2]{\ensuremath \frac{#1}{#2}}
\providecommand\rbrac[1]{\ensuremath \left(#1\right)}

\providecommand\sqbrac[1]{\ensuremath \left[#1\right]}

\providecommand\SQBRAC[1]{\ensuremath \Big[#1 \Big]}
\providecommand\cbrac[1]{\ensuremath \left\{#1\right\}}

\newcommand{\grad}{\nabla}

\theoremstyle{plain}
\newtheorem{theorem}{Theorem}

\theoremstyle{definition}

\newtheorem{remark}[theorem]{Remark}

\let\P\relax
\DeclareMathOperator*{\P}{\mathbb{P}}
\DeclareMathOperator*{\E}{\mathbb{E}}

\definecolor{color_skyblue}{rgb}{0.01,0.39,0.75}

\def \t {\tau}
\def \th {\theta}
\def \a {\alpha}
\def \b {\beta}
\def \p {\varphi}

\def \g {\gamma}

\def \BB {\mathcal{B}}

\def \reals {\mathbb{R}}

%% file: notation.tex
\def \xkp {x_{k+1}}
\def \xk {x_k}

\def \uk {u_k}

\def \xik {\xi_k}

\def \uth {u_\th}
\def \utht {u_{\th^t}}

\def \pith {\pi_\th}
\def \uthp {u_{\th'}}

\def \vth {v^\th}

\def \qp {q_\p}
\def \qpp {q_{\p'}}
\def \qpt {q_{\p^t}}

\def \qpone {q_{\p_1}}
\def \qptwo {q_{\p_2}}

\def \qpot {q_{\p_1^t}}
\def \qptt {q_{\p_2^t}}

\def \td {\textrm{TD}}

\def \d {\textrm{d}}